\documentclass[sigconf]{acmart}
\usepackage[utf8]{inputenc}

\title{Training Sensitivity in Graph Isomorphism Network}

\author{Md. Khaledur Rahman}
\affiliation{%
  \institution{Indiana University Bloomington}}
\email{morahma@iu.edu}

\date{}

\usepackage{natbib}
\usepackage{graphicx}

\begin{abstract}
   Graph neural network (GNN) is a popular tool to learn the lower-dimensional representation of a graph. It facilitates the applicability of machine learning tasks on graphs by incorporating domain-specific features. There are various options for underlying procedures (such as optimization functions, activation functions, etc.) that can be considered in the implementation of GNN. However, most of the existing tools are confined to one approach without any analysis. Thus, this \textcolor{black}{emerging field lacks a robust} implementation ignoring the highly irregular structure of the real-world graphs. In this paper, we attempt to fill this gap by studying various alternative functions for a respective module using a diverse set of benchmark datasets. Our empirical results suggest that the generally used underlying techniques do not always perform well to capture the overall structure from a set of graphs.
\end{abstract}

\copyrightyear{2020}
\acmYear{2020}
\setcopyright{acmcopyright}\acmConference[CIKM '20]{Proceedings of the 29th ACM International Conference on Information and Knowledge Management}{October 19--23, 2020}{Virtual Event, Ireland}
\acmBooktitle{Proceedings of the 29th ACM International Conference on Information and Knowledge Management (CIKM '20), October 19--23, 2020, Virtual Event, Ireland}
\acmPrice{15.00}
\acmDOI{10.1145/3340531.3412089}
\acmISBN{978-1-4503-6859-9/20/10}

\begin{document}
\fancyhead{}


\maketitle
\vspace{-.3cm}
\section{Indtroduction}
\textcolor{black}{Graphs can be considered} as a transformation of knowledge from various domains such as social networks, scientific literature, and protein-interaction networks, where an entity can be denoted by a vertex and the relationship between a pair of entities can be denoted by an edge. An effective representation of a graph can help learn important characteristics from the dataset. For example, a machine learning model can be applied to graph representation to recommend friends in social networks. The traditional graph data structure, such as the adjacency matrix, can be a naive representation for machine learning models for making such predictions. However, we can not accommodate memory for very large graphs as this representation has $O(n^2)$ memory cost. As the size of the representable data in the graph format is growing at a significant speed \cite{reinsel2018digitization}, it \textcolor{black}{allows the researchers} to find an alternate efficient approach. Thus, an effective graph representation learning or embedding technique has become a high demanding problem.

Designing an effective graph embedding method can be extremely challenging as most real-world graphs have a highly irregular structure. Several unsupervised methods have been proposed in the literature to solve this problem sub-optimally \cite{perozzi2014deepwalk,grover2016node2vec}. Graph layout generation methods can also be used to generate embedding \cite{rahman2020batchlayout,rahman2020force2vec}. Some of them generate good quality embedding but have high runtime and memory costs while some of them consume less runtime and memory costs but generate moderate-quality embedding. Nevertheless, these methods can not incorporate domain-specific features which might be helpful to learn a better representation of the graph.
\begin{figure}
    \centering
    \includegraphics[width=\linewidth]{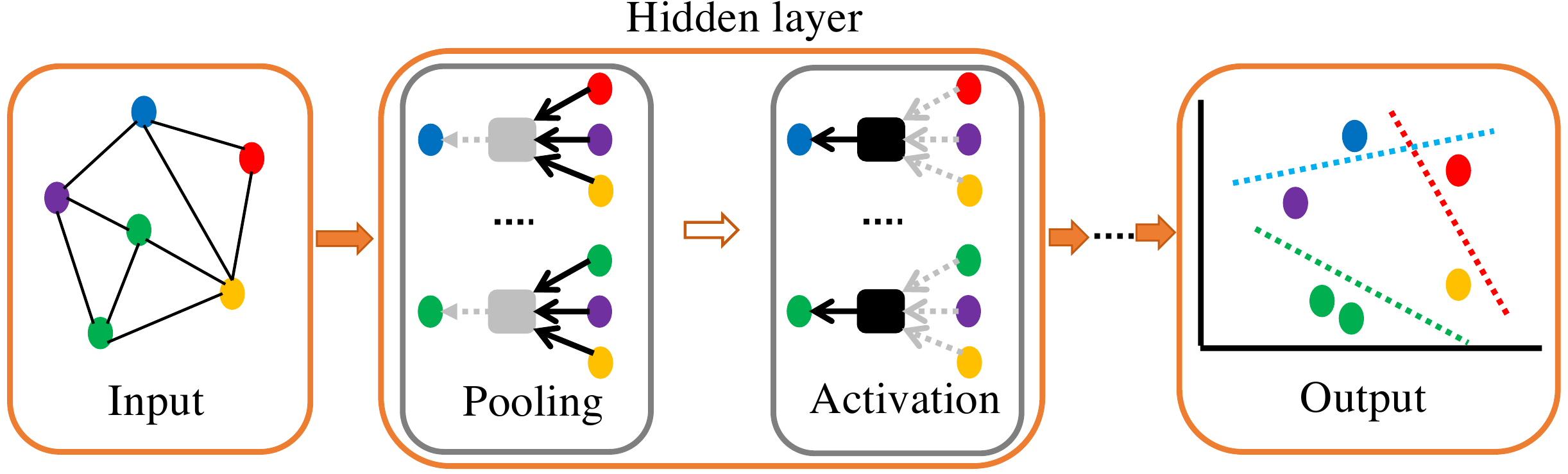}
    \vspace{-0.8cm}
    \caption{A graph neural network with toy example: a hidden layer consists of pooling (aggregation) layer and activation layer. \textcolor{black}{There can be multiple hidden layers in the network}. Output layer represents the embedding of the the network which is generally used for classification task.}
    \label{fig:gcnn}
    \vspace{-0.7cm}
\end{figure}
Recently, Graph Convolutional Network (GCN) has gained much attention in the literature \cite{kipf2016semi}. The basic idea is similar to that of Convolutional Neural Networks (CNN) \cite{krizhevsky2012imagenet} in the computer vision domain where a convolution operator of a vertex aggregates contributions from its neighboring vertices in each hidden layer. A toy example is shown in Fig. \ref{fig:gcnn}. The \emph{GraphSAGE} method further contributes to this field by proposing several effective aggregation functions \cite{hamilton2017inductive}. On the other hand, \emph{FastGCN} runs faster than other methods though it sacrifices the quality of the embedding a little \cite{chen2018fastgcn}. \textcolor{black}{Notably, Errica et al. have performed an extensive set of experiments to compare only existing GNN tools \cite{errica2019fair}}. These types of methods are trained in a semi-supervised manner and used to predict the label of unseen vertices which is often termed as \emph{inductive} learning \cite{hamilton2017inductive}. This is a rapidly growing field; however, the parameter sensitivity of GNN is overlooked in the literature. In this paper, we have filled this gap by analyzing a GNN model, \textcolor{black}{called Graph Isomorphism Network (GIN)~\cite{xu2018powerful}}, for several optimization techniques, aggregation functions, learning rate, and activation functions on a diverse set of benchmark graphs. The summary of our contributions are given as follows:
\begin{itemize}
    \item We develop a set of research questions and \textcolor{black}{address them by parameter tuning to learn the sensitivity of GIN.}
    \item We conduct an extensive set of experiments to analyze the results on a diverse set of benchmark datasets.
    \item Our empirical results provide a new insight that the traditionally used ADAM optimization technique and the ReLU activation function do not always perform better than other alternatives for all types of graphs. This presumably directs us to consider other techniques rather than a single one.
\end{itemize}

\section{Problem Setting}
\subsection{Background}
We represent a graph by $G = (V, E)$, where, $V$ is the set of vertices, $E$ is the set of edges and each $z_i$ of a set $Z$ represents 1-dimensional embedding of $i^{th}$ vertex. We represent neighbors of a vertex $u$ in $G$ by $N(u)$. In a GNN, domain-specific features or one-hot-encoded representation of vertices are fed to the input layer. In \textcolor{black}{a} hidden layer, an aggregation (pooling) function gathers contribution from the neighbors of a vertex to learn the structural properties of the graph that goes through an activation function (see Fig. \ref{fig:gcnn}). Multiple hidden layers can be placed in the network in a cascaded style. The embedding $Z$ of a graph $G$ is produced in the output layer. A cross-entropy based objective function is optimized to update the embedding in a semi-supervised way. \textcolor{black}{The representation of hidden layers in GIN~\cite{xu2018powerful} can be expressed as follows}:
\begin{equation}
\label{eqn:convlayer}
    h^{l}_u = MLP\big ((1+\epsilon^l)h^{l-1}_u+\sum_{v\in N(u)}h^{l-1}_v\big )
    \vspace{-0.3cm}
\end{equation}
Here, $h^l_u$ is the activation of $l^{th}$ hidden layer for vertex $u$ and $h^0_u = X$, where, $X$ is the input features. MLP represents the multi-layer perceptron that helps learn the representation of the network and $\epsilon$ denotes the fraction of considerable activation from the previous layer for the same vertex $u$. A non-linear activation function is used in the MLP of Equation \ref{eqn:convlayer} and a popular choice is Rectified Linear Units (ReLU)~\cite{nair2010rectified}. There are several options to optimize the performance of a GNN by tweaking units such as activation function, optimization technique, or hyper-parameter tuning. 
\vspace{-0.2cm}
\subsection{Activation Functions}
\label{sec:activation}
We employ several activation functions in the GIN, each of which has some benefits over others. We assume that $x$ represents the product of contributions of activation from all neighbors in a hidden layer and the weight matrix of the corresponding layer. Then, the ReLU, LeakyReLU, Sigmoid, and Tanh activation functions compute the activations using the terms $\max(0,x)$, $\max(0,x)+0.01\times\min(0,x)$, $\frac{1}{1+e^{-x}}$, and $\frac{e^{x}-e^{-x}}{e^{x}+e^{-x}}$, respectively. Specifically, we use PyTorch implementations for these activation functions. Generally, the state-of-the-art tools employ one activation function in a GNN such as ReLU in GCN \cite{kipf2016semi} or LeakyReLU in GAT \cite{velivckovic2017graph}, but they do not distinguish among different activation functions. In this paper, we study the above mentioned four standard activation functions.
\vspace{-0.2cm}
\subsection{Optimization techniques}
\label{sec:optimization}
We discuss five techniques to optimize the cross entropy-based loss function in GIN. We study these techniques due to the fact that the graph datasets have irregular structures and one single technique might not perform well to optimize the objective function. 

\textbf{SGD:}
The Stochastic Gradient Descent (SGD) is a widely used algorithm to optimize weights in neural networks \cite{bottou1991stochastic}. This approach updates weight parameters with respect to each training sample which reduces huge computation costs of convergence in the Gradient Descent (GD) approach. 
\vspace{-0.32cm}
\begin{equation}
\label{eqn:sgd}
    W_{t+1} = W_t - \eta \frac{\partial \mathcal{L}}{\partial W_t}
    \vspace{-0.20cm}
\end{equation}
Here, $t$ represents $t^{th}$ iteration, $\eta$ is the learning rate, $\frac{\partial \mathcal{L}}{\partial W_t}$ is the gradient of loss function $\mathcal{L}$ with respect to weight matrix $W$.

\textbf{ADAGRAD:}
The ADAptive GRADient (ADAGRAD) is a variation of the SGD algorithm which focuses on updating multi-dimensional weights using scaled learning rate for each dimension ~\cite{duchi2011adaptive}. We can formally present this updating equation as follows:
\vspace{-0.25cm}
\begin{equation}
\label{eqn:adagrad}
    W_{t+1} = W_t - \frac{\eta}{\sqrt{\epsilon I + diag(G_t)}} \frac{\partial \mathcal{L}}{\partial W_t}
    \vspace{-0.2cm}
\end{equation}
Here, $I$ is the identity matrix, $\epsilon$ is a small value to avoid division by zero, $G_t$ is a diagonal matrix where each diagonal entry represents gradient in the corresponding dimension.

\textbf{ADADELTA:}
ADADELTA is another variation of SGD that solves some of the problems originated by the ADAGRAD approach such as decreasing learning rate over iterations and manual selection of global learning rate~\cite{zeiler2012adadelta}. ADADELTA addresses these issues computing past gradients over a fixed window of steps. To get rid of $\eta$, it also considers accumulated weights similar to gradients. The whole weight update can be represented as follows:
\vspace{-0.25cm}
\begin{subequations}
\begin{align}
    \label{eqn:adadelta}
    U_t(g^2) & = \alpha U_{t-1}(g^2) + (1 - \alpha) g_t^2 \\
    \Delta W_{t} & = \sqrt{\frac{U_{t-1}(\Delta W^2)+\epsilon}{U_t(g^2)+\epsilon}} \\
    U_t(\Delta W^2) & = \alpha U_{t-1}(\Delta W^2) + (1 - \alpha)\Delta W_t^2 \\
    W_{t+1} & = W_t - \Delta W_{t} g_t 
    \vspace{-0.1cm}
\end{align}
\end{subequations}
Here, $g_t$ is gradient of $t^{th}$ iteration, i.e., $g_t = \frac{\partial \mathcal{L}}{\partial W_t}$, $U_t$ is a function that computes accumulation of previous gradients and weights, $\alpha$ is a decaying parameter that determines what percentages of previous gradient and current gradient will be used, respectively.

\textbf{RMSProp:}
RMSProp is similar to ADADELTA with the exception that it does accumulate weight over iterations and it keeps learning rate in the updating equation. This can be formally represented as the following equations:
\vspace{-0.25cm}
\begin{subequations}
\begin{align}
    \label{eqn:rmsprop}
    U_t(g^2) & = 0.9 U_{t-1}(g^2) + 0.1 g_t^2 \\
    W_{t+1} & = W_t - g_t \frac{\eta}{\sqrt{U_t(g^2)+\epsilon}}
    \vspace{-0.2cm}
\end{align}
\end{subequations}

\textbf{ADAM:}
ADAptive Momentum (ADAM) estimation is \textcolor{black}{a} pioneering work that keeps accumulations for the mean and the variance of gradients~\cite{kingma2014adam}. In this approach, a weighted average is calculated by taking a large fraction of gradient from previous steps and a small fraction from the current step. Similarly, the variance of the squared gradient is also taken as a weighted average of past and current squared gradients. We can represent this as follows:
\vspace{-0.25cm}
\begin{subequations}
\begin{align}
    \label{eqn:adam}
    m_t & = \beta_1 m_{t-1} + (1 - \beta_1) g_t \\
    v_t & = \beta_2 v_{t-1} + (1 - \beta_2) g_t^2 \\
    \hat{m_t} & = m_t / (1-\beta^t_1)\\
    \hat{v_t} & = v_t / (1-\beta^t_2)\\
    W_{t+1} & = W_t - \eta \frac{\hat{m_t}}{\sqrt{\hat{v_t}}+\epsilon}
\end{align}
\vspace{-0.1cm}
\end{subequations}
Here, $m_t$ is the momentum of mean gradients and $v_t$ is the momentum of variance of gradients up to $t^{th}$ iteration and $\hat{m_t}$ and $\hat{v_t}$ are bias corrected estimates, respectively. $\beta_1$ and $\beta_2$ are two decaying parameters for the momentum of mean and variance, respectively.

We study the characteristics of graph \textcolor{black}{isomorphism} network using these five types of optimization techniques. 

\subsection{Aggregation/Pooling functions}
\label{sec:aggregation}
We consider three aggregation functions in the GIN architecture \cite{xu2018powerful}. The MAX function aggregates the maximum value from the neighbors. The AVERAGE function sum up contributions from neighbors and aggregates this value dividing by the number of neighbors. The SUM aggregation function simply sums up the contributions from neighbors and send to the activation function. Authors of GIN show that the SUM function can distinguish different structures in the graph whereas other functions fail to do so. Thus, we briefly revisit all these aggregation functions.

\section{Experiments}
\subsection{Experimental Setup}
To conduct experiments, we use a server machine configured as follows: Intel(R) Xeon(R) CPU E5-2670 v3 (2.30GHz), 48 cores arranged in two NUMA sockets, 128GB memory, 32K L1 cache, and 32MB L3 cache. We run all experiments using graph isomorphism network\footnote{\url{https://github.com/weihua916/powerful-gnns/}} which has two MLP layers and generates 64-dimensional embedding of a graph. We set the batch size to 32, number of epoch to 50, and use default values for several hyper-parameters unless otherwise explicitly mentioned. We report the performance of all datasets in terms of accuracy measure i.e., higher the value is better the result is. WLOG, we show the value of the accuracy measure within the range 0 to 1 or multiplying by 100 to report percentage.

\subsection{Datasets}
We use a set of five graph classification datasets which include two bioinformatics networks, and three social networks. We compute the average degree by averaging the avg. degree across all graphs of a particular dataset. This set of graphs have diverse properties such as they have a varying average degree and the number of graphs for each dataset has variable proportions. We provide a summary of the datasets in Table \ref{tab:dataset}. More details about these datasets can be found in \textcolor{black}{the GIN paper} \cite{xu2018powerful}.
\begin{table}[!ht]
\centering
\vspace{-0.2cm}
\caption{Benchmark datasets for graph embedding}
\vspace{-0.25cm}
\begin{tabular}{|c|c|c|c|c|}
\hline
\textbf{Name} & \textbf{\#Graphs} & \textbf{Avg. $|V|$} & \textbf{\#Lab.} & \textbf{Avg. Deg.} \\ \hline
PROTEINS            & 1113              & 39.1           & 2                 &          3.73        \\ \hline
COLLAB        & 5000              & 74.5           & 3                 &      37.38           \\ \hline
NCI1     & 4110             & 29.8         & 2                &              2.15   \\ \hline
REDDITM     & 5000             & 508.5         & 5                &           2.25      \\ \hline
IMDBM     & 1500             & 13.0         & 3                &           8.10      \\ \hline
\end{tabular}
\label{tab:dataset}
\vspace{-0.5cm}
\end{table}

\subsection{Results}
Different graph\textcolor{black}{s} have different structural properties. So, we can presumably say that \textcolor{black}{various} optimization techniques can show \textcolor{black}{dissimilar} results for \textcolor{black}{various graph} datasets. Most of the previously proposed GNNs are \textcolor{black}{lacking} this study. In this paper, we attempt to study these behaviors on several benchmark datasets. Specifically, we employ different optimization techniques, aggregation functions, and activation functions to study the sensitivity of our focused \textcolor{black}{GIN} model. For all experiments, we perform 10-fold cross-validation i.e., we report the results of training accuracy based on 90\% dataset and testing accuracy based on the remaining 10\% dataset. We develop a set of research questions which are discussed as follows.

\begin{figure}[!ht]
    \centering
    \frame{\includegraphics[width=0.49\linewidth]{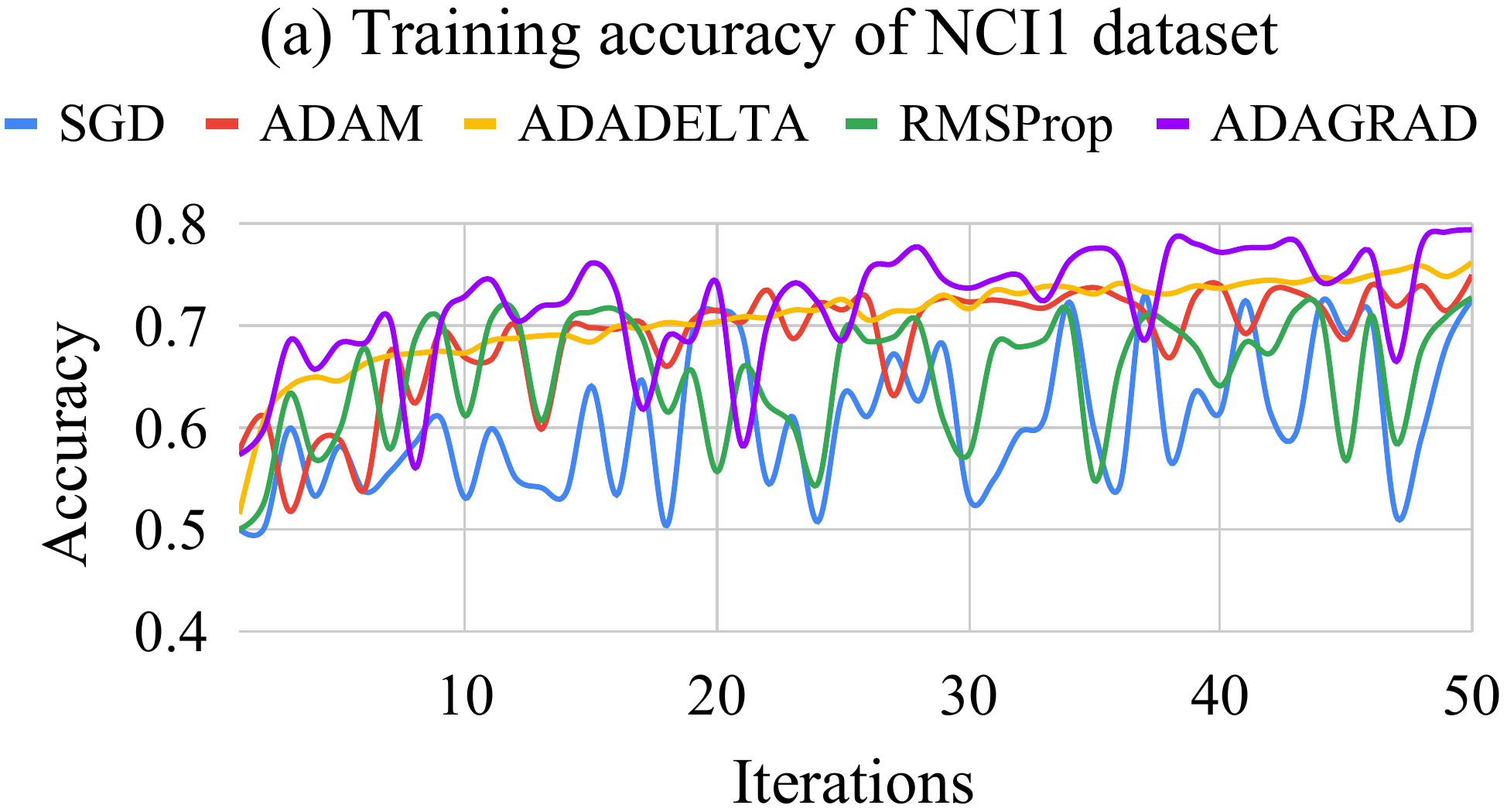}}
    \frame{\includegraphics[width=0.49\linewidth]{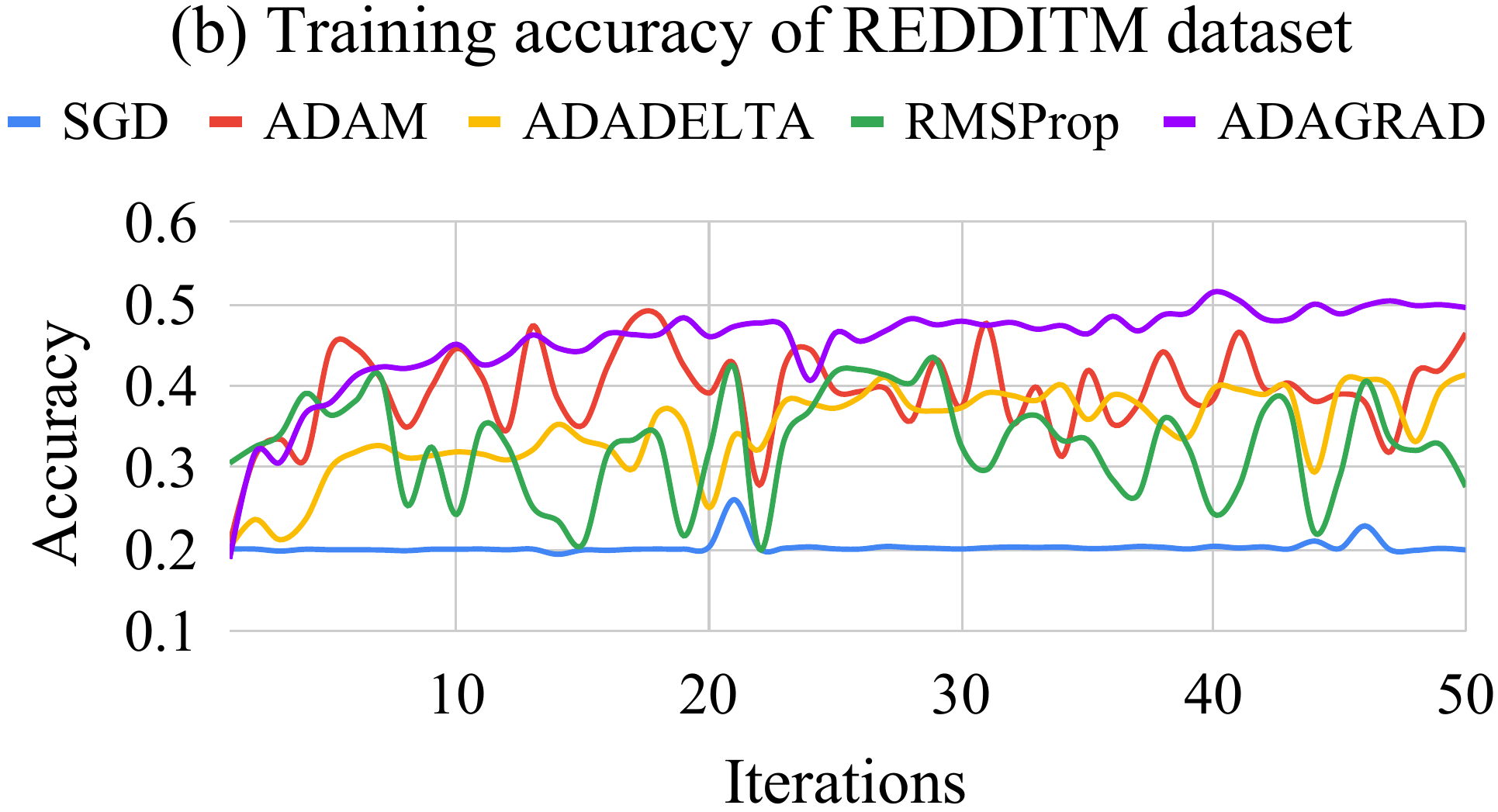}}
    \vspace{-0.2cm}
    \caption{Training accuracy of (a) NCI1 and (b) REDDITM datasets for different optimization techniques.}
    \label{fig:optimizers}
    \vspace{-0.6cm}
\end{figure}

\subsubsection{Which optimization technique is robust for a GNN?}
Most of the GNNs use ADAM optimizer in the programming model~\cite{kipf2016semi,hamilton2017inductive,chen2018fastgcn,xu2018powerful}. So, we measure the performance of different optimization techniques along with ADAM which are discussed in Section \ref{sec:optimization}. We use default values for other parameters in the GIN model. We report the training accuracy of NCI1 and REDDITM datasets in Figs. \ref{fig:optimizers} (a) and (b). We observe that the ADAGRAD optimization technique shows better accuracy curve for different iterations than other optimization techniques. ADADELTA technique also shows better accuracy curve than ADAM. For the REDDITM dataset, ADAGRAD again shows a better performance curve which is more robust than other optimization techniques. So, we can clearly state from the empirical evidence that the ADAGRAD optimization technique is better than ADAM to use in the graph \textcolor{black}{isomorphism} network. We skip training results for other graphs due to space restriction.

\begin{figure*}
    \centering
    {\includegraphics[width=0.24\linewidth,height=2.8cm]{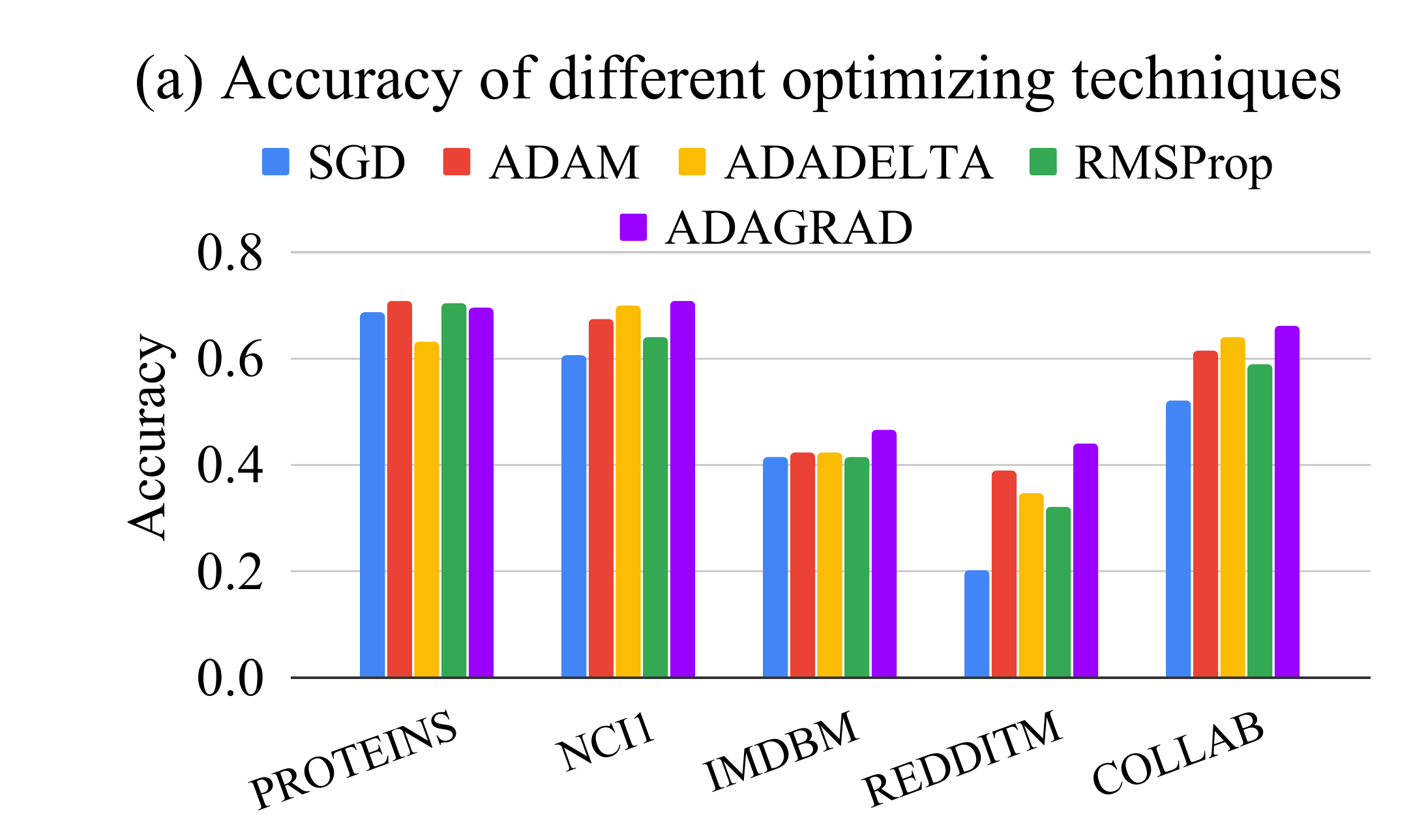}}
    \includegraphics[width=0.24\linewidth,height=2.8cm]{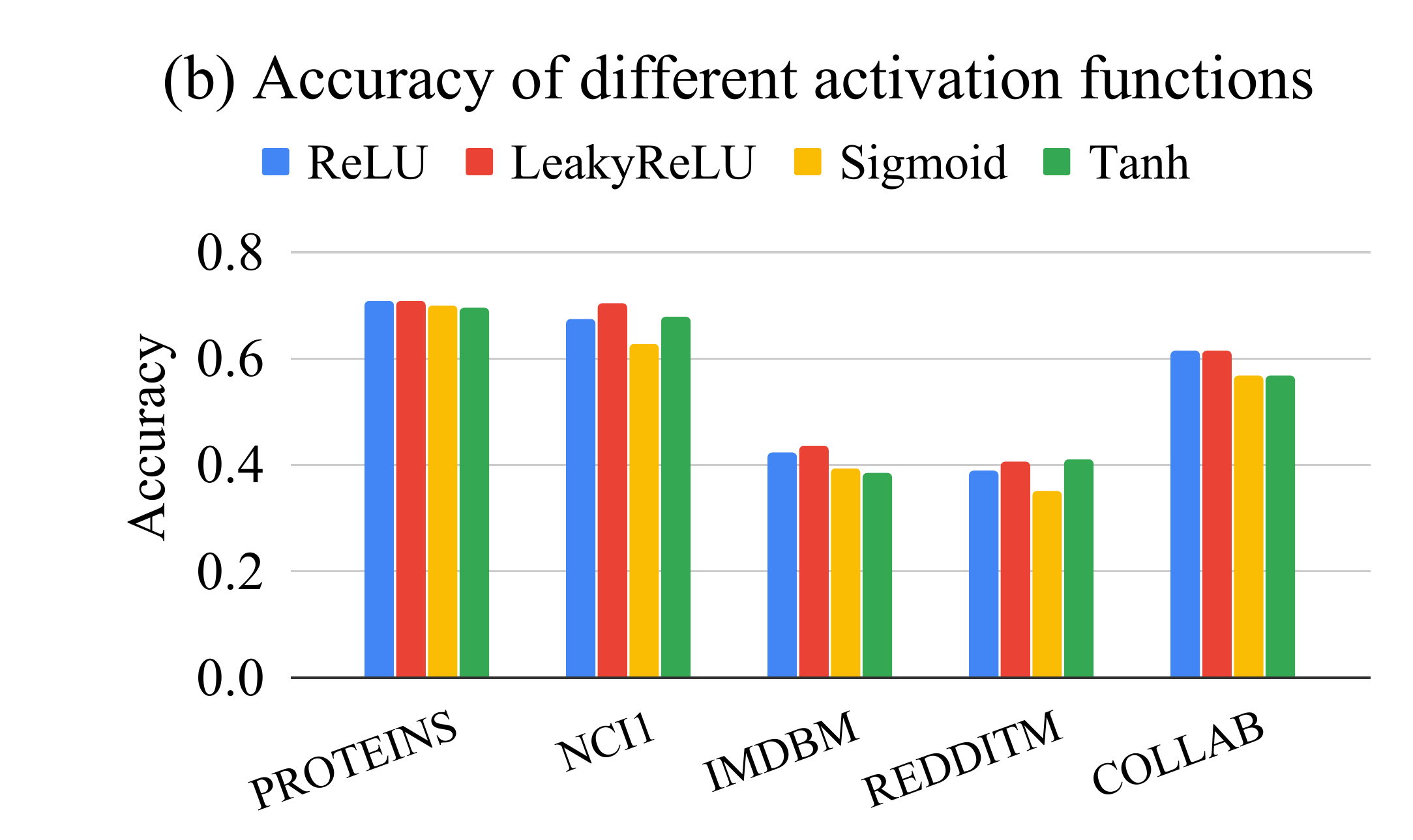}
    \includegraphics[width=0.24\linewidth,height=2.8cm]{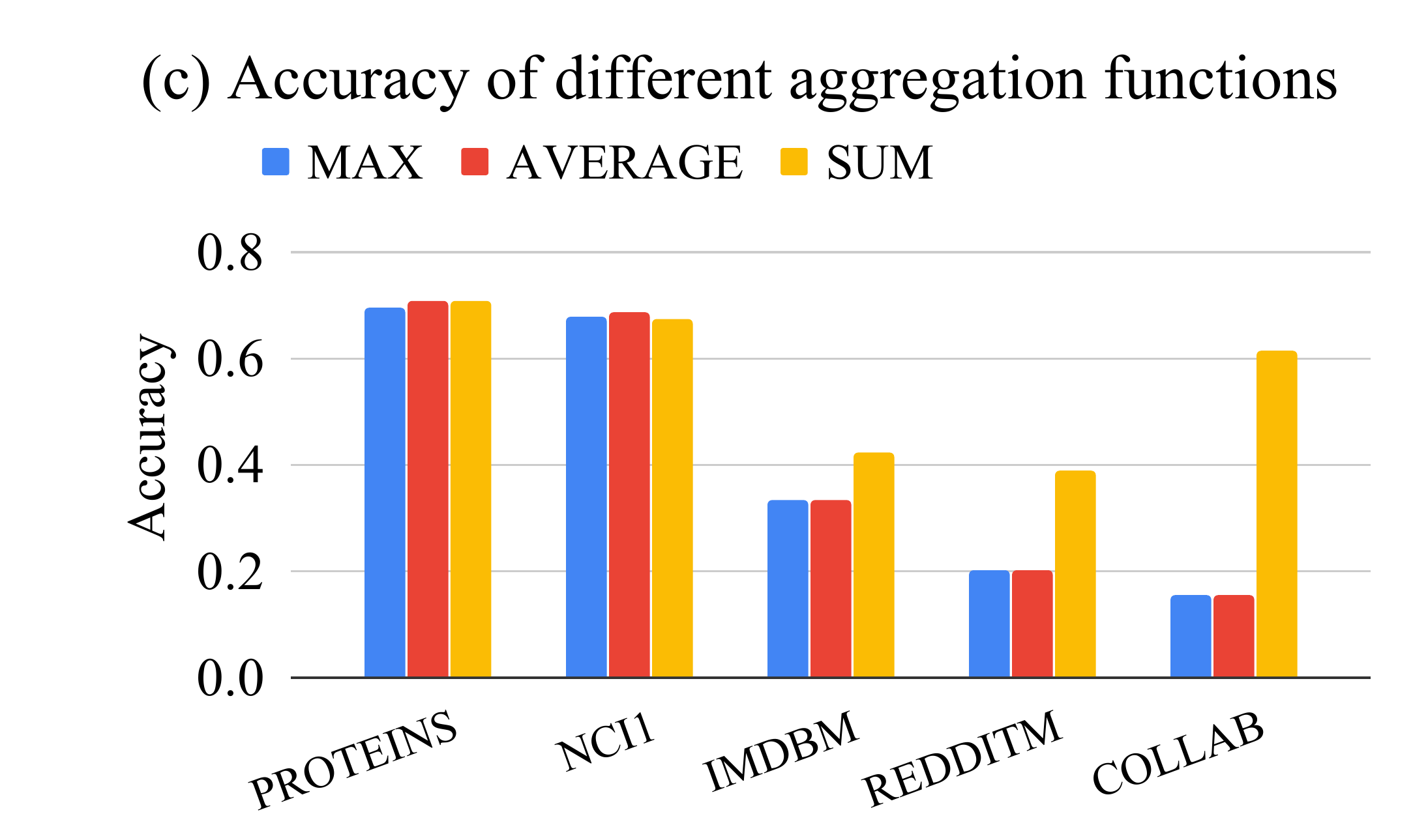}
    \includegraphics[width=0.24\linewidth,height=2.8cm]{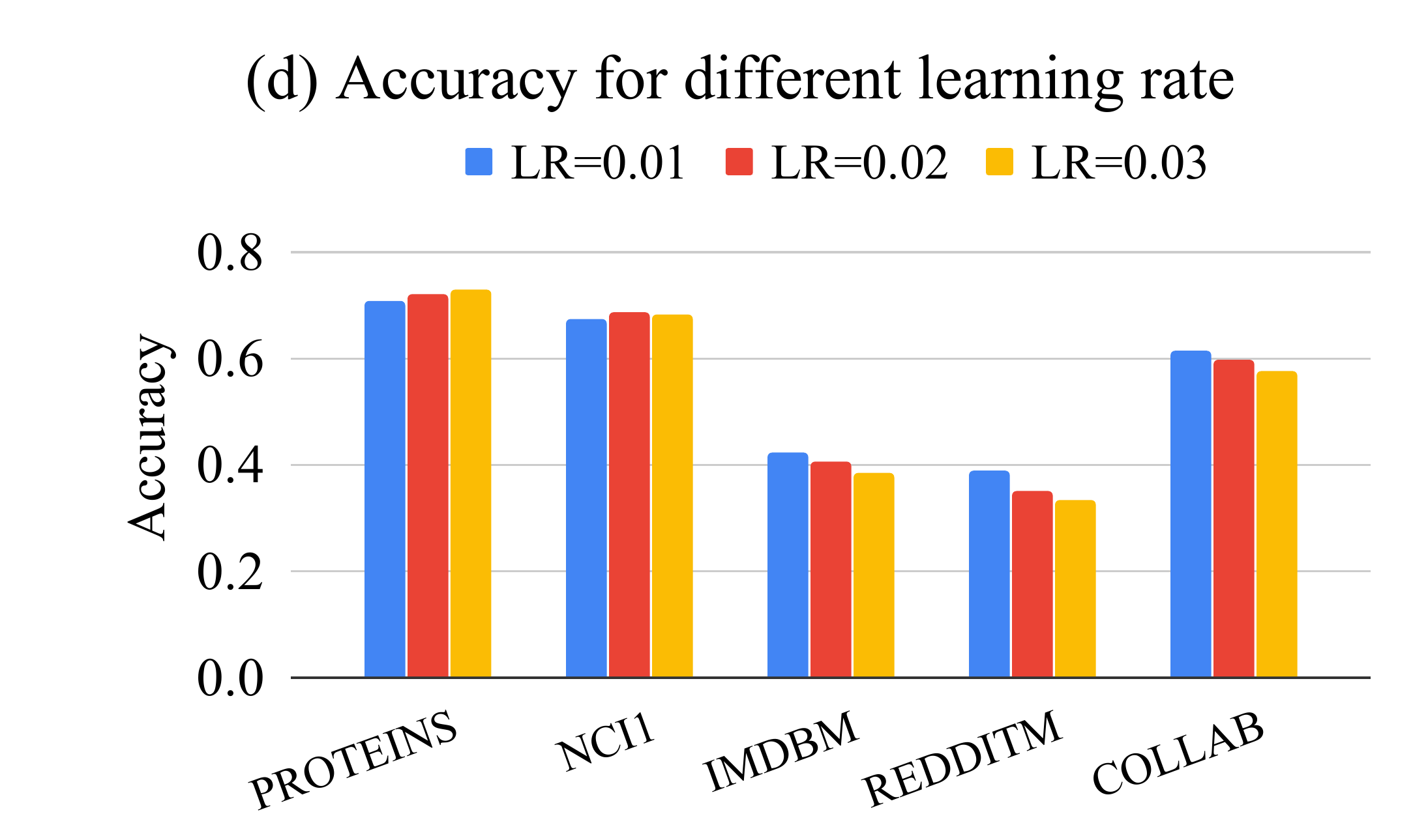}
    \vspace{-0.3cm}
    \caption{Test accuracy of different datasets for different (a) optimization techniques, (b) activation functions, (c) aggregation functions, and (d) learning rates. Average value of test accuracy is reported over several iterations.}
    \label{fig:testaccuracy}
    \vspace{-0.36cm}
\end{figure*}

We show the results of test datasets for different optimization techniques in Fig. \ref{fig:testaccuracy} (a). We observe that the ADAGRAD also performs better than ADAM for NCI1 and REDDITM datasets. In fact, ADAGRAD shows better or competitive performance across all benchmark datasets. \textcolor{black}{We perform a two-sided Wilcoxon signed-rank test on 50 samples of test results which also supports our findings ($p$-value < 0.017) for all datasets \cite{wilcoxon1992individual}}. This empirical evidence gives us a new insight to use the ADAGRAD technique \textcolor{black}{instead of ADAM in graph neural networks}.  

\begin{figure}[!ht]
    \centering
    \frame{\includegraphics[width=0.49\linewidth]{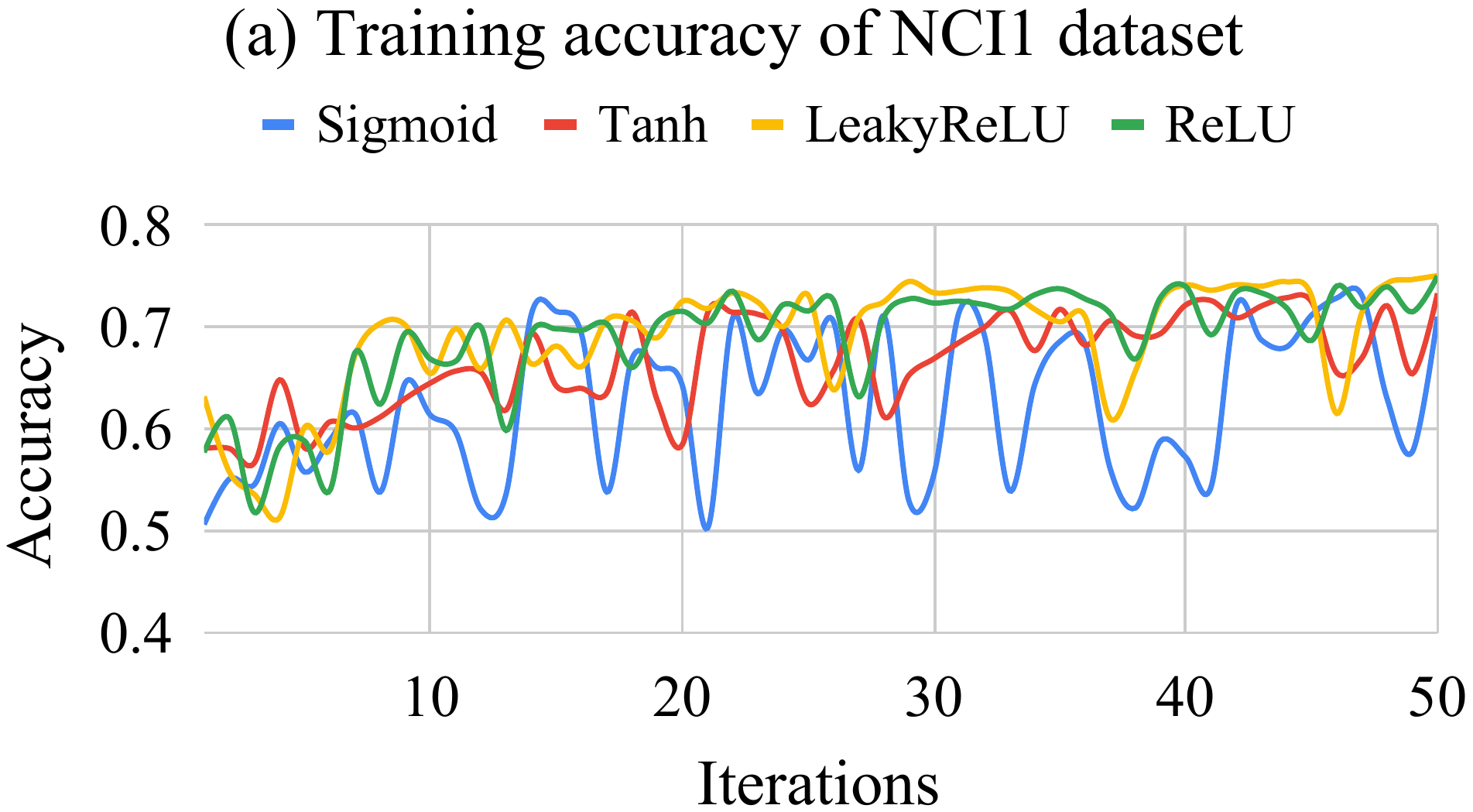}}
    \frame{\includegraphics[width=0.49\linewidth]{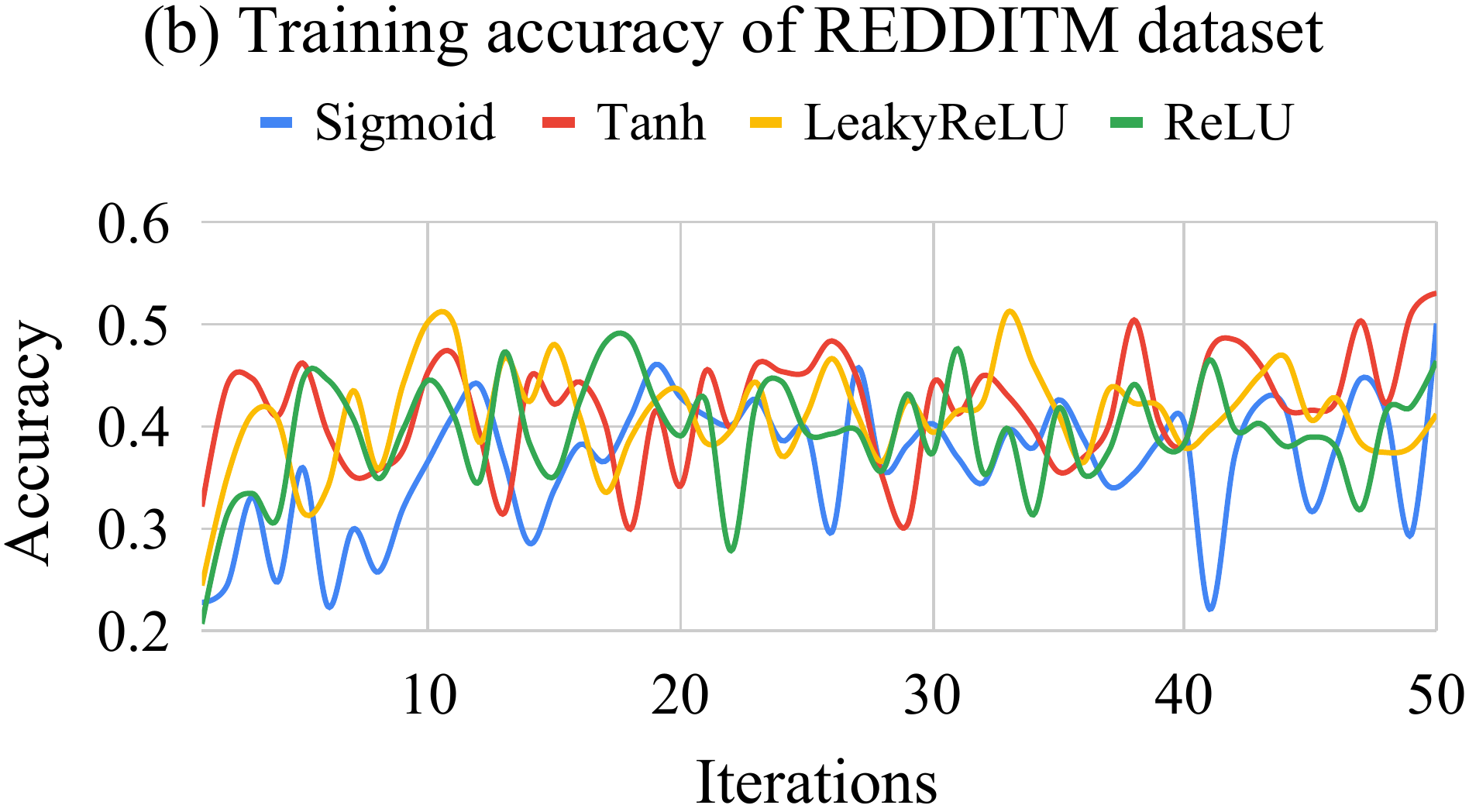}}
    \vspace{-0.2cm}
    \caption{Training accuracy of (a) NCI1 and (b) REDDITM datasets for different activation functions.}
    \label{fig:activations}
    \vspace{-0.5cm}
\end{figure}

\subsubsection{Which activation function is more effective?}
Similar to ADAM optimization technique, most existing graph neural networks use ReLU activation function in the programming model \cite{kipf2016semi,hamilton2017inductive,xu2018powerful}. So, we evaluate the performance of other activation functions along with ReLU which are discussed in Section \ref{sec:activation}. As usual, we use default values for other parameters in the model. We report the results of training accuracy for NCI1 and REDDITM datasets in Figs. \ref{fig:activations} (a) and (b). The training accuracy for all activation functions shows spikes in the learning curves. However, we observe that LeakyReLU and ReLU shows competitive performance over several iterations while training the model. For the REDDITM dataset, the Tanh activation function also shows competitive performance.

We report the results for test datasets for all activation functions in Fig. \ref{fig:testaccuracy} (b). We observe that LeakyReLU shows better or competitive performance than other activation functions. Notice that the Tanh activation technique shows better test accuracy than ReLU for REDDITM dataset. These experimental results provide us a new insight that the LeakyReLU might be a better choice for activation function in graph neural networks than ReLU.

\subsubsection{Aggregation functions revisited}
Authors of the GIN method demonstrate several scenarios where MAX and AVERAGE aggregation functions can fail to distinguish between structural differences of graphs. On the other hand, the SUM aggregation function can correctly distinguish such structural differences. Our experimental results also support this statement. We briefly report the results of the test datasets in Fig. \ref{fig:testaccuracy} (c). We observe that all aggregation functions perform almost equally well for PROTEINS and NCI1 datasets. However, the SUM function performs significantly better than other functions for IMDBM, REDDITM, and COLLAB datasets. Thus, SUM would be an order invariant powerful aggregation function for graph neural networks.
\begin{figure}[!ht]
    \centering
    \vspace{-0.25cm}
    \frame{\includegraphics[width=0.49\linewidth]{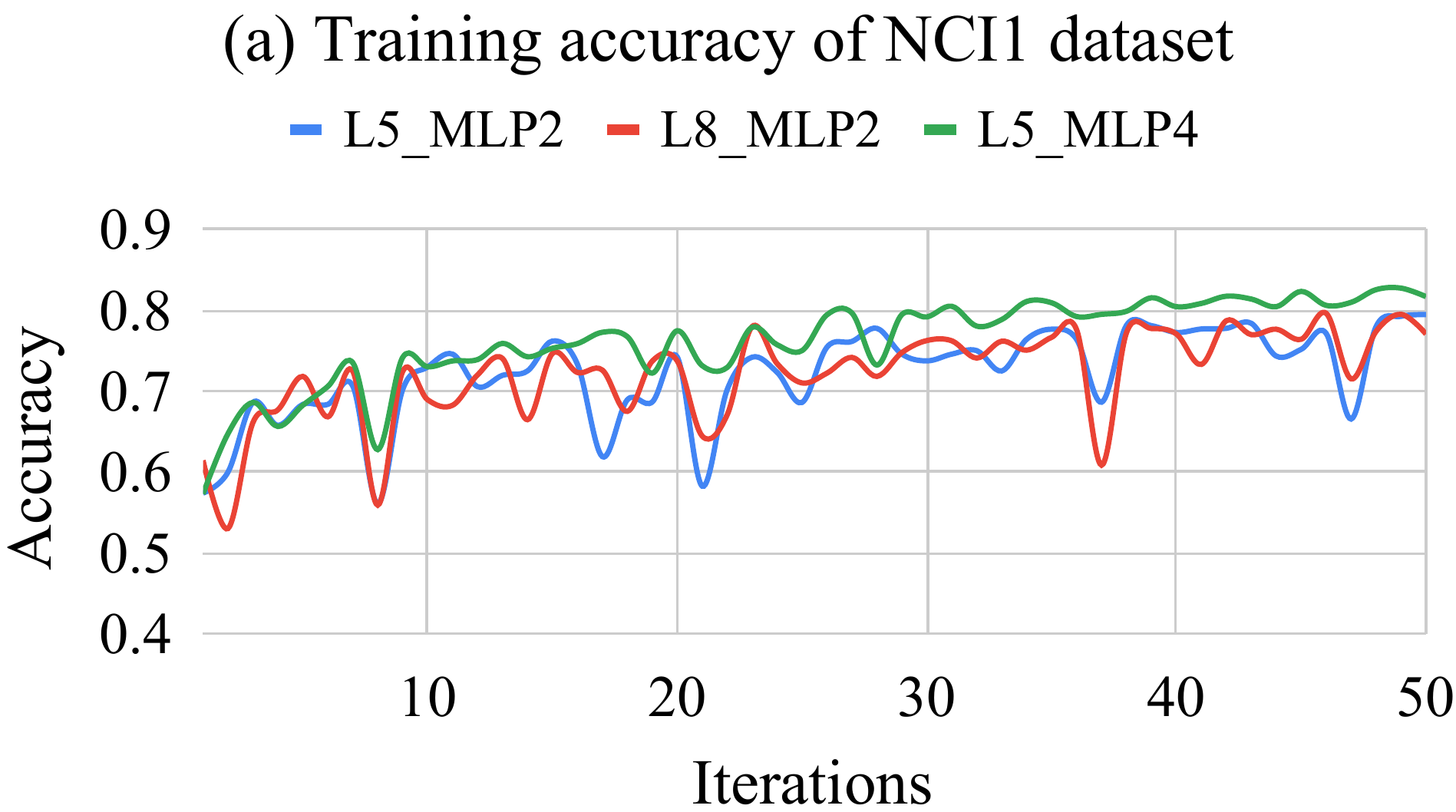}}
    \frame{\includegraphics[width=0.49\linewidth]{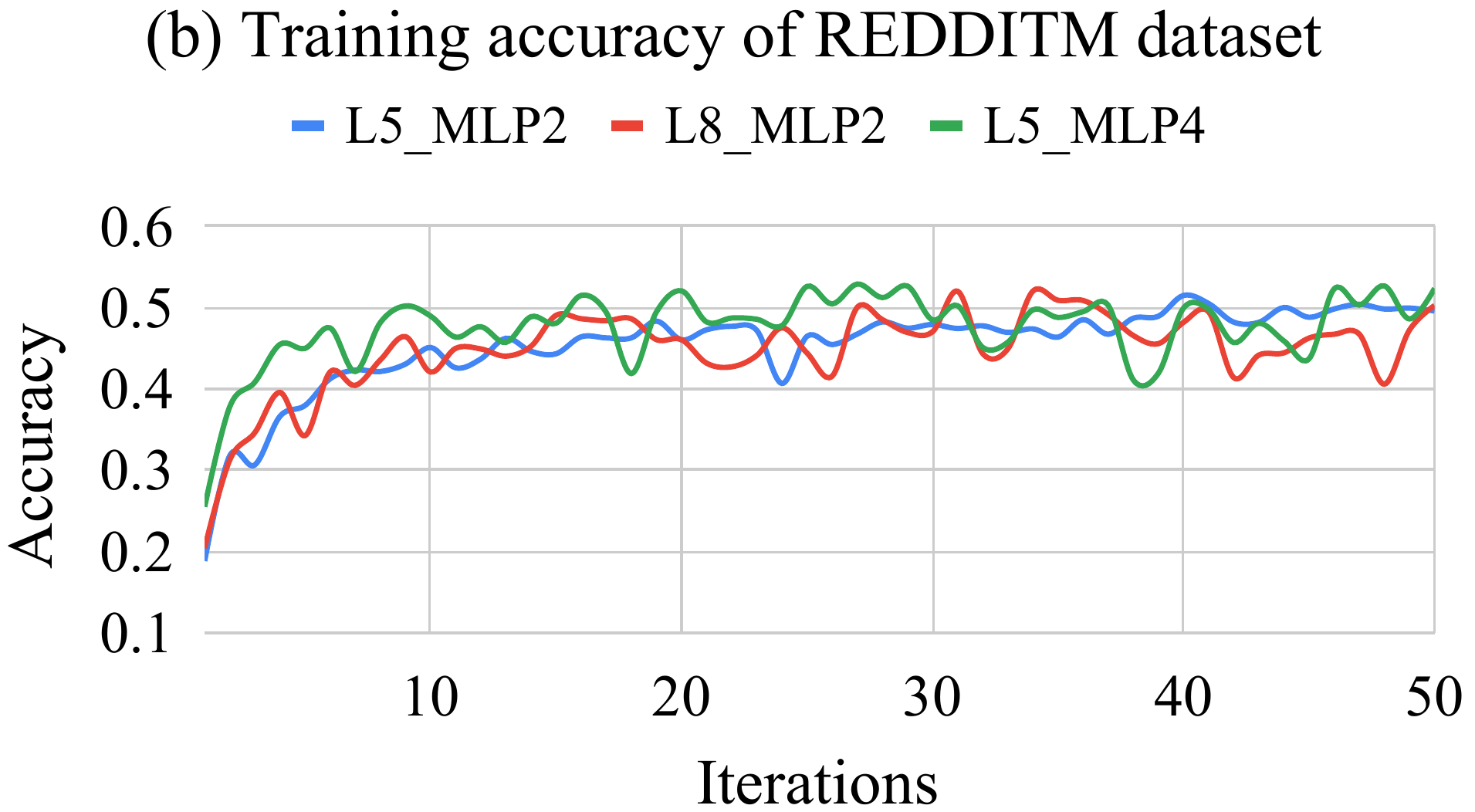}}
    \vspace{-0.2cm}
    \caption{Training accuracy of (a) NCI1 and (b) REDDITM datasets for different number of hidden and MLP layers.}
    \label{fig:mlplayer}
    \vspace{-0.6cm}
\end{figure}
\subsubsection{Hyper-parameter sensitivity}
We also study several hyper-parameters in graph \textcolor{black}{isomorphism} network such as hidden dimension, learning rate, the number of hidden layers, and the number of layers in MLP. Notably, we use the default value for other parameters in the model. The parameter sensitivity of the embedding dimension has been studied in supervised learning \cite{perozzi2014deepwalk,grover2016node2vec}. So, we also conduct experiments varying the size of the hidden dimension in GIN for the REDDITM dataset. For this experimental setup, we get a test accuracy of 39.1\%, 39\%, 39\%, and  38.4\% for 16, 32, 64, and 128-dimensional embedding, respectively. Thus, we do not see any significant difference for other datasets as well but in general, accuracy may drop for very higher-dimensional representation.

We report the results of test accuracy for different learning rates in Fig. \ref{fig:testaccuracy} (d). We observe that a learning rate of 0.02 shows better results for PROTEINS and NCI1 datasets whereas a learning rate of 0.01 shows better performance for IMDBM, REDDITM, and COLLAB datasets. Thus, a learning rate of 0.01 would be a reasonable choice for the GIN model. We show training accuracy in Figs. \ref{fig:mlplayer} (a) and (b) for NCI1 and REDDITM datasets, respectively, varying the number hidden layer and MLP layer. In both figures, L5\_MLP2 means that the learning model has 5 hidden layers and 2 MLP layers. We observe that increasing the number of hidden layers in the model does not change the performance significantly; however, increasing the number of MLP layers contributes to improve the performance. So, a higher value for MLP layers might be a better option to seek improved performance.

\vspace{-0.1cm}
\section{Conclusions}
Graph neural network has become a powerful toolbox for analyzing and mining different graph datasets. However, this field is \textcolor{black}{lacking} a rigorous evaluation of several underlying techniques that can be employed in GNN. For example, most existing methods stick to a particular optimization technique such as ADAM or an activation function such as ReLU. In this paper, we empirically demonstrate that ADAGRAD and LeakyReLU might be better options for optimization technique and activation function, respectively. We believe that our findings would provide a new insight into the community for choosing better underlying techniques while developing a new graph neural network.
\vspace{-0.1cm}

\bibliographystyle{acm}
\bibliography{references}

\begin{thebibliography}{10}

\bibitem{bottou1991stochastic}
{\sc Bottou, L.}
\newblock Stochastic gradient learning in neural networks.
\newblock {\em Proceedings of Neuro-N{\i}mes 91}, 8 (1991), 12.

\bibitem{chen2018fastgcn}
{\sc Chen, J., Ma, T., and Xiao, C.}
\newblock {FastGCN}: fast learning with graph convolutional networks via
  importance sampling.
\newblock {\em arXiv:1801.10247\/} (2018).

\bibitem{duchi2011adaptive}
{\sc Duchi, J., Hazan, E., and Singer, Y.}
\newblock Adaptive subgradient methods for online learning and stochastic
  optimization.
\newblock {\em Journal of machine learning research 12}, Jul (2011),
  2121--2159.

\bibitem{errica2019fair}
{\sc Errica, F., Podda, M., Bacciu, D., and Micheli, A.}
\newblock A fair comparison of graph neural networks for graph classification.
\newblock {\em arXiv:1912.09893\/} (2019).

\bibitem{grover2016node2vec}
{\sc Grover, A., and Leskovec, J.}
\newblock node2vec: Scalable feature learning for networks.
\newblock In {\em KDD\/} (2016), ACM, pp.~855--864.

\bibitem{hamilton2017inductive}
{\sc Hamilton, W., Ying, Z., and Leskovec, J.}
\newblock Inductive representation learning on large graphs.
\newblock In {\em NIPS\/} (2017), pp.~1024--1034.

\bibitem{kingma2014adam}
{\sc Kingma, D.~P., and Ba, J.}
\newblock Adam: A method for stochastic optimization.
\newblock {\em arXiv preprint arXiv:1412.6980\/} (2014).

\bibitem{kipf2016semi}
{\sc Kipf, T.~N., and Welling, M.}
\newblock Semi-supervised classification with graph convolutional networks.
\newblock {\em arXiv preprint arXiv:1609.02907\/} (2016).

\bibitem{krizhevsky2012imagenet}
{\sc Krizhevsky, A., Sutskever, I., and Hinton, G.~E.}
\newblock Imagenet classification with deep convolutional neural networks.
\newblock In {\em NIPS\/} (2012), pp.~1097--1105.

\bibitem{nair2010rectified}
{\sc Nair, V., and Hinton, G.~E.}
\newblock Rectified linear units improve restricted boltzmann machines.
\newblock In {\em ICML\/} (2010), pp.~807--814.

\bibitem{perozzi2014deepwalk}
{\sc Perozzi, B., Al-Rfou, R., and Skiena, S.}
\newblock {DeepWalk}: Online learning of social representations.
\newblock In {\em KDD\/} (2014), ACM, pp.~701--710.

\bibitem{rahman2020batchlayout}
{\sc Rahman, M.~K., Sujon, M.~H., and Azad, A.}
\newblock {BatchLayout}: A batch-parallel force-directed graph layout algorithm
  in shared memory.
\newblock In {\em 2020 IEEE Pacific Visualization Symposium (PacificVis)\/}
  (2020), IEEE, pp.~16--25.

\bibitem{rahman2020force2vec}
{\sc Rahman, M.~K., Sujon, M.~H., and Azad, A.}
\newblock {Force2Vec: Parallel force-directed graph embedding}.
\newblock {\em In Review\/} (2020).

\bibitem{reinsel2018digitization}
{\sc Reinsel, D., Gantz, J., and Rydning, J.}
\newblock The digitization of the world from edge to core.
\newblock {\em IDC White Paper\/} (2018).

\bibitem{velivckovic2017graph}
{\sc Veli{\v{c}}kovi{\'c}, P., Cucurull, G., Casanova, A., Romero, A., Lio, P.,
  and Bengio, Y.}
\newblock Graph attention networks.
\newblock {\em arXiv:1710.10903\/} (2017).

\bibitem{wilcoxon1992individual}
{\sc Wilcoxon, F.}
\newblock Individual comparisons by ranking methods.
\newblock In {\em Breakthroughs in statistics}. Springer, 1992, pp.~196--202.

\bibitem{xu2018powerful}
{\sc Xu, K., Hu, W., Leskovec, J., and Jegelka, S.}
\newblock How powerful are graph neural networks?
\newblock {\em arXiv preprint arXiv:1810.00826\/} (2018).

\bibitem{zeiler2012adadelta}
{\sc Zeiler, M.~D.}
\newblock Adadelta: an adaptive learning rate method.
\newblock {\em arXiv:1212.5701\/} (2012).

\end{thebibliography}
\end{document}